\def\year{2022}\relax
\documentclass[letterpaper]{article} 
\usepackage{aaai22}  
\usepackage{times}  
\usepackage{helvet}  
\usepackage{courier}  
\usepackage[hyphens]{url}  
\usepackage{graphicx} 
\usepackage{natbib}  
\usepackage{caption} 
\DeclareCaptionStyle{ruled}{labelfont=normalfont,labelsep=colon,strut=off} 
\usepackage{algorithm}
\usepackage{algorithmic}
\usepackage{multirow}
\usepackage{booktabs}
\usepackage{hhline}
\newcommand{\ra}[1]{\renewcommand{\arraystretch}{#1}}

\usepackage{newfloat}
\usepackage{listings}
\floatstyle{ruled}
\newfloat{listing}{tb}{lst}{}
\floatname{listing}{Listing}
\usepackage{todonotes}

\title{Automatic Evaluation and Moderation of Open-domain
Dialogue Systems}
\author{
    Chen Zhang\textsuperscript{\rm 1}, 
    João Sedoc\textsuperscript{\rm 2}, 
    Luis Fernando D'Haro\textsuperscript{\rm 3}, 
    Rafael Banchs\textsuperscript{\rm 4}, 
    Alexander Rudnicky\textsuperscript{\rm 5}
}
\affiliations{
    \textsuperscript{\rm 1}National University of Singapore, Singapore \\
    \textsuperscript{\rm 2}Department of Technology, Operations, and Statistics, New York University, USA \\
    \textsuperscript{\rm 3}Speech Tecnology and Machine Learning Group - Universidad Politécnica de Madrid, Spain\\
    \textsuperscript{\rm 4}Intapp Inc., USA \\
    \textsuperscript{\rm 5}Carnegie Mellon University, USA\\

}

\usepackage{bibentry}

\begin{document}

\maketitle

\begin{abstract}
The development of Open-Domain Dialogue Systems (ODS) is a trending topic due to the large number of research challenges, large societal and business impact, and advances in the underlying technology. However, the development of these kinds of systems requires two important characteristics: 1) automatic evaluation mechanisms that show high correlations with human judgements across multiple dialogue evaluation aspects (with explainable features for providing constructive and explicit feedback on the quality of generative models' responses for quick development and deployment) and 2) mechanisms that can help to control  chatbot responses, while avoiding toxicity and employing intelligent ways to handle toxic user comments and keeping interaction flow and engagement. 
This track at the 10\textsuperscript{th} Dialogue System Technology Challenge (DSTC10) is part of the ongoing effort to promote scalable and toxic-free ODS. This paper describes the datasets and baselines provided to participants, as well as submission evaluation results for each of the two proposed subtasks. 

\end{abstract}

\section{Introduction}
In recent years, dialogue systems have attracted significant academic and industry interest. Especially the discipline of open-domain dialogue systems, aka chatbots, which has gained great momentum. Yet, a long-standing challenge that concerns researchers is the lack of effective automatic evaluation metrics, which results in a significant research impediment \cite{yeh2021comprehensive}. Common practice in assessing the performance of open-domain dialogue models involves extensive human evaluation of the final deployed models, which is both time- and cost- intensive. During the model development phase, researchers must rely on standard automatic metrics, such as BLEU \cite{papineni2002bleu} and
perplexity, to tune the performance of their models. However, these metrics correlate poorly with human judgements \cite{liu2016not} resulting in suboptimal dialogue systems.

Moreover, a recent trend in building open-domain chatbots involves pretraining dialogue models with a large amount of social media conversation data \cite{zhang2019dialogpt,adiwardana2020towards, smith2020can}. However, the interaction data from social media conversations may include offensive and inappropriate content. Indiscriminate usage of such data can result in insensitive and toxic generative models. Recently, \citep{xu2020recipes} proposes recipes for safety in open-domain chatbots, such as unsafe utterance detection and safe utterance generation. Although these recipes help provide safe responses to toxic comments, the safe chatbot tends to avoid directly responding to these comments by switching to other topics. Sometimes, simply ignoring such comments may not be enough. Especially in the domain of customer support where customer service personnel must answer occasional offensive complaints in a polite and appropriate way.

This paper provides a comprehensive overview of Track 5 on ``Automatic Evaluation and Moderation of Open-domain Dialogue Systems" organized as part of the 10\textsuperscript{th} Dialogue System Technology Challenge (DSTC10). The paper is structured as follows. In section \ref{section:task1}, we completely describe our first subtask on automatic evaluation metrics, including datasets, baseline, and participants' results. Then, in section \ref{section:task2}, we describe the datasets, baselines, and subjective and objective evaluation metrics for our second subtask on safe chatbot development. Finally, section \ref{section:conclusions} presents our main conclusions and directions for future work.   

\begin{table*}[!t]
\centering
\resizebox{\textwidth}{!}{
    \begin{tabular}{l|cccccccl}
    \toprule
    Name & \#Instances & Avg.\#Utts. & Avg.\#Ctx/Hyp Words & Type & \#Criteria & \#Annotations & Used NLG models\\
    \midrule
    Persona-USR~\shortcite{mehri-eskenazi-2020-usr} & 300 & 9.3 & 98.4 / 12.0  & Turn & 6 & 5,400 & Transformer Seq2Seq, LSTM LM, Memory Network\\
    ConvAI2-GRADE~\shortcite{huang-etal-2020-grade} & 600 & 3.0 & 24.4 / 11.3 & Turn & 1 & 3,000 & Transformer Seq2Seq, DialoGPT, BERT/Transformer Ranker  \\
    Persona-Zhao~\shortcite{zhao-etal-2020-designing} & 900 & 5.1 & 48.8 / 11.5 & Turn & 1 & 3,600 & LSTM Seq2Seq, and GPT-2\\
    DailyDialog-GRADE~\shortcite{huang-etal-2020-grade} & 300 & 3.0 & 26.0 / 10.8 & Turn & 1 & 3,000 & Transformer Seq2Seq, Transformer Ranker\\
    DailyDialog-Zhao~\shortcite{zhao-etal-2020-designing} & 900 & 4.7 & 47.5 / 11.0 & Turn & 4 & 14,400 & LSTM Seq2Seq, Random, and GPT-2\\
    DailyDialog-Gupta~\shortcite{gupta-etal-2019-investigating} & 500 & 4.92 & 49.9 / 10.9 & Turn & 1 & 2,500 & LSTM Seq2Seq, Conditional VAE\\
    Topical-USR~\shortcite{mehri-eskenazi-2020-usr} & 360 & 11.2 & 236.3 / 22.4 & Turn & 6 & 6,480 & Transformers\\
    Empathetic-GRADE~\shortcite{huang-etal-2020-grade} & 300 & 3.0 & 29.0 / 15.6 & Turn & 1 & 3,000 & Transformer Seq2Seq, Transformer Ranker \\
    Reddit-DSTC7~\shortcite{Galley2019GroundedRG} & 9,990 & 3.5 & 35.3 / 11.2 & Turn & 3 & 29,700 & RNN, LSTM Seq2Seq, Memory Network, Pointer-generator\\
    Twitter-DSTC6~\shortcite{hori2017end} & 40,000 & 2.0 & 27.74 / 20.77 & Turn & 1 & 400,000 & LSTM Seq2Seq Variants \\
    FED-Turn~\shortcite{mehri-eskenazi-2020-unsupervised}  & 375 & 10.4 & 87.3 / 13.3 & Turn & 9 & 16,863 & Meena, Mitsuku\\
    HUMOD~\shortcite{app10030762} & 9,500 & 3.9 & 17.0 / 6.1 & Turn & 2 & 57,000 & Random sampling\\
    \midrule
    FED-Dial~\shortcite{mehri-eskenazi-2020-unsupervised}  & 125 & 12.7 & 113.8 / - & Dialogue & 11 & 6,720 & Meena, Mitsuku\\
    Persona-See~\shortcite{see-etal-2019-makes}  & 3316 & 12.0 & 91.07 / - & Dialogue & 9 & 29,844 & LSTM Seq2Seq with Different Controlling Strategies\\
    \bottomrule
    \end{tabular}
}
    \caption{Summary of the development datasets. Some information are from~\citep{yeh2021comprehensive}.}
    \label{tab:eval-data}
\end{table*}

\section{Automatic Evaluation Metrics}
\label{section:task1}

The goal of this subtask is for participants to design robust automatic dialogue evaluation metrics that correlate well with human judgements across multiple dialogue domains as well as across different dialogue evaluation dimensions, such as naturalness, appropriateness, etc. We allow the participants to use any existing resources (open-source human-human dialogue datasets, pretrained models, existing metrics, etc) for designing their own model-based evaluation metrics. There is one exception: the participants are not allowed to directly train supervised models on datasets containing human-annotated quality scores. The reason is that such models can easily overfit to the training datasets and hence, lose generalizability for performing evaluation of dialogues in new domains.

\subsection{Datasets}
\label{subsec:adm-dataset}
During the system development phase, an evaluation benchmark, which consists of 14 publicly available datasets, was released for the participants to tune the performance of their proposed metrics. During the final evaluation phase, we have collected five hidden test evaluation datasets for assessing participants' submissions.

\subsubsection{Development Datasets}
The detailed statistics of the 14 development evaluation datasets are outlined in Table~\ref{tab:eval-data} and each dataset is outlined as follows:

\textbf{The GRADE Datasets }
\citet{huang-etal-2020-grade} collected three evaluation datasets, Empathetic-GRADE, DailyDialog-GRADE and ConvAI2-GRADE, which are collected based on dialogues in the test sets of EmpatheticDialogues~\citep{rashkin-etal-2019-towards}, DailyDialog~\citep{li2017dailydialog} and ConvAI2\footnote{Dev set of ConvAI2}~\citep{dinan2020second} respectively. Each context-response pair is annotated by 8-10 different AMT turkers. The turkers are asked to assess the coherence between a context and the corresponding response on a scale of 1-5 (not coherent at all to very coherent). Since only the human scores for each pair are publicly available and there is no information regarding the annotators, we assume that the same group of annotators consistently annotated all context-response pairs. Hence, the inter-annotator agreements of Empathetic-GRADE, DailyDialog-GRADE and ConvAI2-GRADE in terms of Spearman correlations are 0.376, 0.423 and 0.453 respectively.


\textbf{DailyDialog-Zhao}~\citep{zhao-etal-2020-designing} evaluation dataset is collected based on 100 dialogues sampled from the test set of the DailyDialog corpus~\citep{li-etal-2017-dailydialog}. In DailyDialog-Zhao, four criteria are assessed: appropriateness, language usage, relevance, and content. Each context-response pair is rated by four annotators on a 5-point Likert scale. The Krippendorff's $\alpha$ along appropriateness after removal of outliers is above 0.8.



\textbf{DailyDialog-Gupta}~\citep{gupta-etal-2019-investigating} is constructed based on 100 dialogue contexts from the test set of DailyDialog. In DailyDialog-Gupta, each context-response pair is annotated by 5 different AMT workers along the appropriateness dimension (from 1-5). According to the original paper, ratings of annotators with a Cohen’s Kappa inter-annotator agreement of less than 0.2 are removed. The remaining workers have a mean kappa of 0.43, indicating moderate agreement.


\textbf{Persona-Zhao}~\citep{zhao-etal-2020-designing} evaluation dataset is constructed in a similar manner as DailyDialog-Eval. The context-response pairs of Persona-Zhao are collected based on dialogues from the test set of the PersonaChat corpus~\citep{zhang2018personalizing}. Only the appropriateness quality of the response is annotated in Persona-Zhao, with an inter-annotator agreement above 0.8 in terms of Krippendorff's $\alpha$.

\textbf{The USR Datasets }
\citet{mehri-eskenazi-2020-usr} developed two high-quality human evaluation datasets, Topical-USR and Persona-USR. The same annotation schemes are applied to both datasets. Each context-response pair is annotated by three dialogue researchers along six different dialogue quality categories: Understandable (0-1), Natural (1-3), Maintains Context (1-3), Interesting (1-3), Uses Knowledge (0-1), Overall Quality (1-5)\footnote{The numbers in the bracket are the Likert scales.}. The inter-annotator agreements for the above six annotation categories of USR-Topical are: 0.5102, 0.4871, 0.5599, 0.5811, 0.7090, and 0.7183 respectively in terms of Spearman correlation scores. For USR-Persona, the inter-annotator agreements of the six annotation categories are: 0.2984, 0.4842, 0.6125, 0.4318, 0.8115 and 0.6577 respectively.


\textbf{HUMOD}~\citep{app10030762} is a high-quality human annotated multi-turn movie dialogue dataset developed from the Cornell Movie--Dialogs Corpus~\citep{danescu2011chameleons}. HUMOD contains human annotations on fictional conversations of the movie scripts and diverse human generated replies. Each context-response pair in HUMOD is annotated by three different annotators. The annotators provide scores between 1 and 5 to indicate the degree of relevance between a response w.r.t. the corresponding context. The inter-annotator agreements of HUMOD along the relevance criteria are 0.836 and 0.836 respectively, in terms of Pearson and Spearman correlations.


\textbf{Twitter-DSTC6}~\citep{hori2017end} is the largest among all evaluation datasets (40000 context-response pairs). Each context-response pair in Twitter-DSTC6 is annotated by 10 different Turkers using 5-point Likert Scale. The annotation is based on whether the responses are relevant to the respective dialogue context. The inter-annotator agreement of Twitter-Eval is 0.421 and 0.476, respectively, in terms of Pearson and Spearman correlations.


\textbf{Reddit-DSTC7}~\citep{Galley2019GroundedRG} consists of knowledge-grounded conversations. For each  context-response pair, three crowd-sourced annotators provide scores based on two criteria, relevance and informativeness. The scores for each criterion are based on the 5-point Likert scale. The overall score is obtained by combining the two judgments with equal weights.


\textbf{Persona-See}~\citep{see-etal-2019-makes} evaluation dataset contains 3,316 conversations from 26 model configurations including a human agent. The annotation is performed at the dialogue level in an interactive fashion. An annotator chats with one model configuration for 6 conversational exchanges. At the end of the conversation, the annotator rates the interaction based on eight criteria: avoiding repetition, interestingness, making sense, fluency, listening, inquisitiveness, humanness and engagingness. All questions use a 1-4 Likert scale\footnote{Except for avoiding repetition and inquisitiveness}, the higher, the better. On average, there are 114 conversations per model configuration and each model configuration has been evaluated by over 100 annotators. 

\textbf{FED}~\citep{mehri-eskenazi-2020-unsupervised} consists of 124 conversations: 40 come from Meena, 44 come from Mitsuku and another 40 are drawn from human-human conversations. Quality annotations are performed at both the dialogue level and turn level. There were 9 turn-level criteria, and 11 dialog-level criteria. We denote the turn- and dialogue-level evaluation datasets FED-Dial and FED-Turn respectively. The inter-annotator agreements of FED-Turn and FED-Dial along each evaluation criteria range from approximately 0.70 to 0.85 in terms of Spearman correlations, indicating high agreement\footnote{Except understandability for turn-level, and consistency for dialogue-level}.

\subsubsection{Test Datasets}

\subsubsection{CHANEL-JSALT-2020 (JSALT)}
The JSALT dataset includes validity annotations on a 3 point-scale of dialog segments \citep{kong2019subjective} from the EmpatheticDialogues and the TopicalChat datasets. They take each dialog segment and have it annotated by four different annotators. This dataset consists of only human continuations of the dialogues.

\subsubsection{ChatEval Datasets -- Neural Conversation Model (NCM) \& English As a Second Language (ESL)}
We also evaluate against several datasets released by
\citet{sedoc-etal-2019-chateval} including the Neural Conversational Model (NCM) and ESL three turn dialogue segment datasets. The NCM dataset is a collection of hand-crafted 200 single turn prompts developed  by \citet{vinyals2015neural}. The 200 ESL dialogue segments are from an English learning website.\footnote{\url{https://www.eslfast.com/}} NCM and ESL datasets contain pairwise comparisons between system responses.
NCM has 59 comparisons between 11 systems and 2 human baselines with at least 3 annotators for each prompt. The dataset has over 33K pairwise comparisons.
ESL has 21 comparisons of 5 systems and a human baseline with just over 13K judgements~\citep{lee2020evaluation}.
We compute the win ratio for each human reference-model response pair and normalize by the number of comparisons. 
The win ratio most closely represents the \textit{Overall Quality} since it captures human preference between two candidate responses. 



\subsubsection{DSTC10-T5.1 Evaluation Set (Topical-DTSC10 \& Persona-DSTC10)}
As part of DSTC10 Track 5, we create a new dataset. We use the framework provided by \citet{zhao-etal-2020-designing}. Our only change to the survey is that we include 8 systems, a human baseline, and a random utterance instead of 3 at a time. Specifically, the 8 systems are LSTM Seq2Seq, attention-based LSTM Seq2Seq, HRED, VHRED, BlenderBot (400M-Distill)~\citep{roller-etal-2021-recipes}, DialoGPT (medium)~\citep{zhang2019dialogpt}, T5 (base)~\citep{raffel2020exploring}, and GPT-3~\citep{brown2020language}. They cover a wide quality spectrum of dialogue systems. 

Our dataset can be divided into two sub-datasets based on the domains. We denote them Topical-DTSC10 and Persona-DSTC10. For both datasets, we sample 500 dialogue segments from the conversations in the test set of TopicalChat and PersonaChat, respectively. In total, we collected 4500 context-response pairs (9 responses per context) for Topical-DTSC10 and 5000 context-response pairs (10 responses per context) for Persona-DSTC10. Each context-response pair is rated by four annotators following \citet{zhao-etal-2020-designing}. After applying mean average deviation filtering, the annotator agreement as measured by Krippendorff's alpha is 0.688.


%

\subsection{Baseline}

We adopt the deep AM-FM framework~\citep{zhang-etal-2021-deepamfm}, an ensemble metric, as the baseline for the automatic dialogue evaluation task\footnote{\url{https://github.com/e0397123/dstc10_metric_track}}. We modify the framework to a reference-free version whereby for AM, we compute the cosine similarity between the sentence-level embedding of the response and that of the last sentence in the corresponding dialogue context. For FM, we follow the formulation of the context-response coherence metric in HolisticEval~\citep{pang-etal-2020-towards}. Motivated by~\citep{zhang2021investigating}, we choose RoBERTa-base~\citep{liu2019roberta} as the backbone pretrained language model of AM. We further adapt the pretrained model to a combination of four dialogue corpora: DailyDialog~\citep{li-etal-2017-dailydialog}, TopicalChat~\citep{gopalakrishnan2019topical}, ConvAI2~\citep{dinan2020second}, and EmpatheticDialogues~\citep{rashkin-etal-2019-towards} with the mask language modeling objective. The backbone of FM is a GPT2-medium model~\citep{gpt2} that has adapted to the same above-mentioned combination of dialogue corpora with the causal language modeling objective.



\subsection{Participants}

In total, we received 21 and 35 submissions from nine different teams for development and testing, respectively. We request each team to provide a short system description w.r.t. their submissions. Below is the list of descriptions collected from the participants:

\subsubsection{Team 1 System Description}

Team 1 experimented with a broad range of ideas, ranging from a single DynaEval model~\citep{DBLP:conf/acl/ZhangCDZFL020} to the ensemble of multiple metrics. Team 1 improves DyanEval's performance on turn-level evaluation by adding auxiliary objectives such as next sentence prediction, and response selection. In their ensemble approach, team 1 combines USL-H~\citep{phy-etal-2020-deconstruct}, DEB~\citep{DBLP:journals/tacl/SaiMAK20}, and the improved DyanEval metric with weights determined by the characteristics of input dialogue data.  

\subsubsection{Team 4 System Description}
Inspired by a recent work on characterizing Twitter SpamBots as humans~\citep{giorgi2021characterizing}, team 4 employs five human-centered metrics, including \emph{emotional entropy}, \emph{linguistic style matching}, \emph{emotion matching}, \emph{agreeableness}, and \emph{empathy}. These metrics are proposed based on the assumption that dialogues are part of a psychologically grounded hierarchical process.

\subsubsection{Team 5 System Description}
Team 5 proposes an ensemble metric consisting of 5 metric categories with 7 distinct sub-metrics, to holistically evaluate the quality of dialogues. A novel score composition method, Correlation Re-Scaling (CRS), is adopted to model the relationship between the sub-metrics and various dialogue qualities.


\subsubsection{Team 8 System Description} 
Team 8 proposes a framework named \textit{IM}$^{2}$ (\textit{I}nterpretable and \textit{M}ulti-category \textit{I}ntegrated \textit{M}etric) to tackle the multi-dimensional, and multi-datasets automatic dialogue evaluation task. Firstly, team 8 groups a list of evaluation metrics into four categories with each target one aspect of the dialogues, specifically, FI-Metric for first impression, NUF-metric for response quality, CR-metric for context relevance, and IES-Metric for specificity. The scores w.r.t. each category are combined with linear regression to derive the final \textit{IM}$^{2}$ metric score.

\subsubsection{Team 9 System Description}
For turn-level evaluation datasets, team 9 employs two QuantiDCE~\citep{DBLP:conf/acl/YeLHLL20} variants: (1) QuantiDCE model pretrained on the DailyDialog++ dataset~\citep{DBLP:journals/tacl/SaiMAK20}. (2) QuantiDCE modelc finetuned with the respective evaluation datasets via knowledge distillation. For dialogue-level evaluation datasets, DynaEval~\citep{DBLP:conf/acl/ZhangCDZFL020} is adopted for correlation analysis.

\subsection{Results}
Table~\ref{tab:correlation-main-turn} presents the main correlation results of each team on the five test datasets. For each row in the table, we show the Spearman rank correlation w.r.t. each team's best submission. Each entry at row 6 is computed by averaging the 11 dimension-wise correlation scores over all the five datasets. Each dimension-wise correlation score is computed between the metric scores assigned to all the data instances within a test dataset and the corresponding human annotated scores along one evaluation criteria of that particular dataset.

Based on the results in row 6, Teams 5, 8, and 1 rank first, second,  and third, respectively.
Team 5 performs the best on Topical-DSTC10 and Persona-DSTC10. Team 1 performs the best on JSALT and NCM. Team 9 performs the best on ESL. Remarkably, Team 1, 5, and 8 all rely on ensembling multiple sub-metrics for evaluation. The weights of combining different sub-metrics are dynamically learnt from the data. This finding is inline with the observation made in~\citet{yeh2021comprehensive}, which highlights the advantage of combining multiple sub-metrics.

Table~\ref{tab:correlation-dev-results} presents the correlation results of each team on the 14 development datasets. It can be observed that Team 7 performs exceptionally well with an average correlation score of 52.15\%, outperforming the second best team by a large margin of around 13\%, and achieving the best performance on 11 datasets. Team 8 and Team 6 rank the second, and the third respectively. 

In general, all teams' performance on the test datasets is worse compared to that on the development datasets (Table~\ref{tab:correlation-dev-results}). Surprisingly, the performance of Team 7 on the test datasets is significantly worse compared to their performance on the development datasets. All teams' performance drop is expected as the test datasets and development datasets are of different distributions. This not only showcases that the test datasets are challenging, but also highlights the need to continue developing robust metrics that can generalize to unseen evaluation datasets. 

\begin{figure}[!t]
    \centering
    \includegraphics[width=0.5\textwidth]{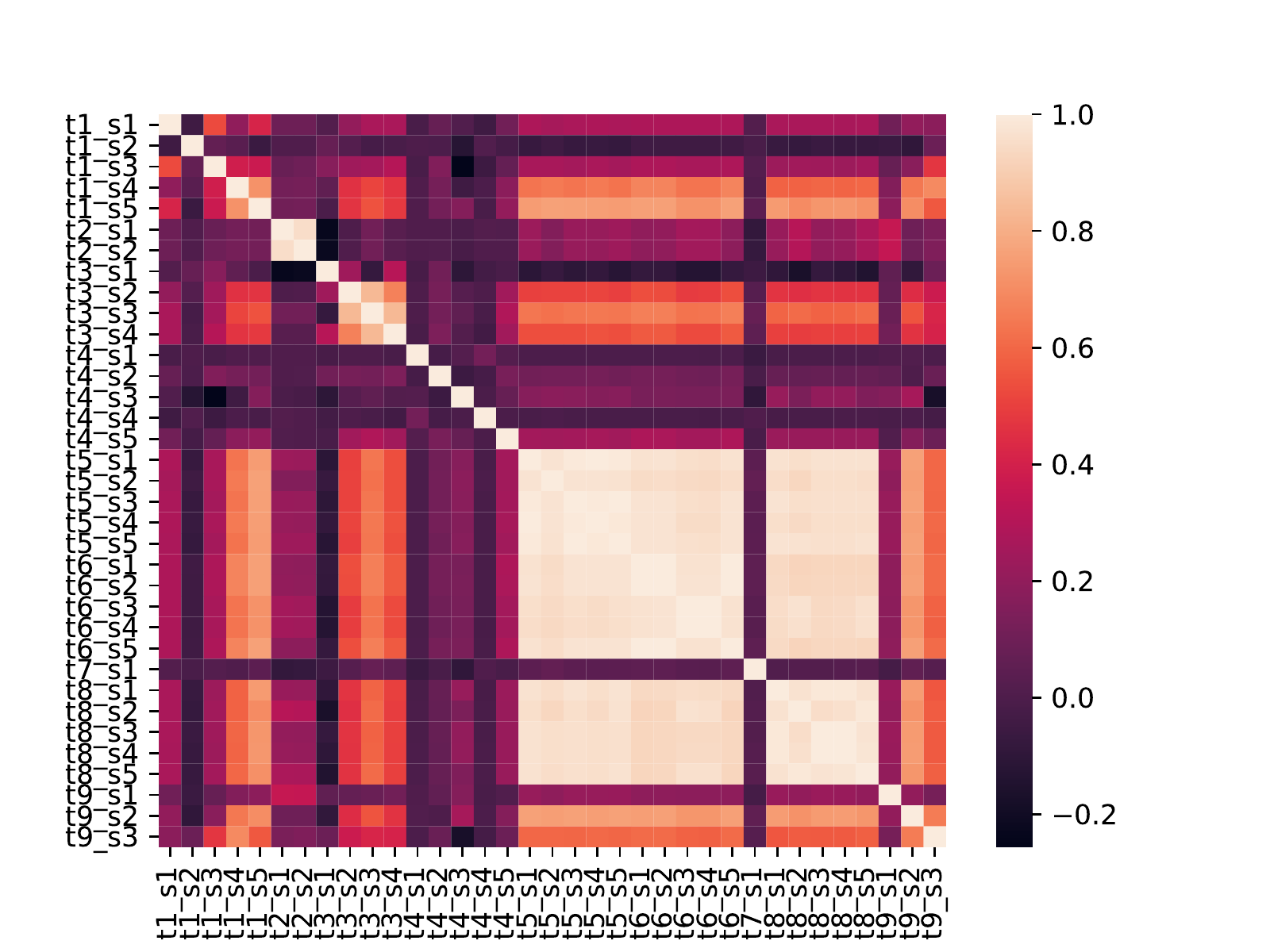}
    \caption{Spearman correlations w.r.t. different pairs of test submissions. The name of each submission is denoted as t$\{x\}$\_s$\{y\}$ where $x$ is the team number and $y$ is the submission number.}
    \label{fig:pairwise_corr}
\end{figure}

Figure~\ref{fig:pairwise_corr} demonstrates the pairwise Spearman correlation of all 35 submissions. Each submission contains 18,641 metric scores w.r.t test instances of all five test datasets. Interestingly, we can observe clusters (Teams 5,6, and 8), thus indicating effectively similar approaches. However, some teams submitted quite different metrics even within the team (e.g., Teams 1 and 4). This points out that there may be value in ensembling these metrics.

\begin{table*}[!ht]	
\centering
\small
\resizebox{\linewidth}{!}{
\begin{tabular}{@{}|c|l|cccccccccc|@{}}
\toprule
\textbf{Row} & \textbf{Datasets}  & \textbf{Baseline} & \textbf{Team 1} & \textbf{Team 2} & \textbf{Team 3} & \textbf{Team 4} &  \textbf{Team 5} & \textbf{Team 6} & \textbf{Team 7} & \textbf{Team 8} & \textbf{Team 9}\\ \midrule
1 & JSALT & 5.09 & \textbf{27.74} & 3.10 & 10.54 & 4.96 & 11.66 & \emph{12.73} & 4.07 & 8.75 & \underline{26.42} \\
2 & ESL & 32.29 & \underline{43.18} & 19.86 & 28.75 & 9.34 & \emph{40.01} & 32.92 & 3.28 & 36.10 & \textbf{45.58} \\
3 & NCM & 16.49 & \textbf{29.91} & 1.98  & 22.08 & 8.24 & \underline{29.60} & \emph{26.60} & 2.01 & 25.57 & 19.11 \\
4 & Topical-DSTC10 & 17.48 & \emph{21.32} & 10.85 & 14.56 & 8.33 & \textbf{23.68} & 20.00 & 1.43 & \underline{22.77} & 17.41 \\
5 & Persona-DSTC10 & 19.61 & 30.67 & 7.77 & 25.80 & 16.59 & \textbf{37.50} & \emph{35.78} & 2.54 & \underline{37.22} & 33.82 \\
\midrule
6 & Average & 18.38 & \emph{27.81} & 8.95 & 20.20 & 10.29 & \textbf{29.63} & 26.86 & 2.30 & \underline{28.19} & 26.89 \\  \bottomrule
\end{tabular}}
\caption{Average Spearman correlations (\%) of the baseline as well as the best submission from each team on 5 test datasets. The best score for each row is highlighted in bold. The second best is underlined. The third best is italicized. Note that each entry at row 6 is averaged over 11 dimension-wise correlation scores of all five datasets instead of over the entries of rows 1-5 in the same column.}
\label{tab:correlation-main-turn}
\end{table*}

\begin{table*}[!ht]	
\centering
\small
\resizebox{\linewidth}{!}{
\begin{tabular}{@{}|c|l|cccccccccc|@{}}
\toprule
\textbf{Row} & \textbf{Datasets}  & \textbf{Baseline} & \textbf{Team 1} & \textbf{Team 2} & \textbf{Team 3} & \textbf{Team 4} &  \textbf{Team 5} & \textbf{Team 6} & \textbf{Team 7} & \textbf{Team 8} & \textbf{Team 9}\\ \midrule
1 & ConvAI2-GRADE & 9.38 & 50.43 & 7.23 & 41.01 & 17.31 & \emph{58.43} & \underline{60.42} & 57.00 & \textbf{60.43} & 53.07\\
2 & DailyDialog-GRADE & 15.48 & \emph{36.30} & 3.45 & 11.16 & 19.86 & 33.42 & 30.00 & \textbf{64.42} & 30.06 & \underline{41.89}\\
3 & DailyDialog-Gupta & 17.70 & 56.78 & 2.76  & 38.16  & 11.39 & \underline{63.25} & \emph{61.37} & \textbf{78.85} & 60.84 & 46.69\\
4 & DailyDialog-Zhao & 22.25 & 36.94 & 20.96 & 33.09 & 19.85 & 48.03 & \underline{52.99} & \textbf{54.50} & \emph{52.79} & 28.70\\
5 & Twitter-DSTC6 & 9.96 & \emph{24.46} & 8.05  & \underline{47.95}  & 4.26 & 17.94 & 18.35 & \textbf{61.63} & 18.31 & 18.54 \\
6 & Reddit-DSTC7  & 2.67 & 33.97 & 19.78  & 25.75 & 12.14 & 32.48 & \textbf{34.15} & 31.30 & \underline{34.12} & \emph{33.16} \\ 
7 & Empathetic-GRADE & 2.51 & \underline{39.52} & 6.38 & 22.59 & 4.70 & 30.57 & 24.62 & \textbf{50.10} & 24.65 & \emph{36.50}\\ 
8 & FED-Turn  & 5.09 & 23.85 & 9.49 & 11.96 & 19.27 & 30.38 & \underline{33.01} & \textbf{35.15} & \emph{32.88}  & 19.87\\ 
9 & HUMOD & 11.73 & 32.86 & 1.93 & 31.11 & 4.16 & \emph{33.20} & \textbf{33.83} & 22.45 & \textbf{33.83} & 22.28 \\ 
10 & Persona-USR  & 14.42 & 27.25 & 12.22 & 21.61 & 26.69 & \underline{40.36} & 35.51 & \textbf{47.88} & \emph{36.17} & 22.60 \\ 
11 & Persona-Zhao  & 46.79& 55.21 & 24.23 & 50.19 & 5.23  & 61.32 & \emph{64.24} & \textbf{76.40} & \underline{64.58} & 55.70 \\ 
12 & Topical-USR & 14.10 & 21.84 & 29.59 & 17.06 & 27.79 & \emph{39.08} & 38.68 & \textbf{45.49} & \underline{40.24} & 13.73 \\
13 & FED-Dial  & 11.18 & 26.92 & 25.22  & 5.70 & 5.93 & 46.89 & \underline{49.31} & \textbf{77.42} & \underline{49.31} & 40.26\\
14 & Persona-See & 8.08 & 5.70 & 3.50 & 6.95 & 3.69 & 8.78 & \underline{12.92} & \textbf{27.52} & \underline{12.92} & 6.27 \\ 
\midrule
15 & Average & 13.67& 33.72 & 12.48 & 26.02 & 13.02 & 38.87 & \emph{39.24} & \textbf{52.15} & \underline{39.37} & 31.38 \\  \bottomrule
\end{tabular}}
\caption{Average Spearman correlations (\%) of the baseline as well as the best submission from each team on 14 development evaluation datasets. The best score for each row is highlighted in bold. The second best is underlined. The third best is italicized.}
\label{tab:correlation-dev-results}
\end{table*}

\section{Safe Chatbot Development}
\label{section:task2}
The goal of this subtask is for participants to build generative models that first detect a toxic user's comment, and then generate appropriate and polite responses that keep the dialogue fluid and nontoxic.

In the literature, we find different definitions of toxicity and related terms such as offensive, hateful, abusive, insulting, etc. \cite{fortuna2020toxic}. For this task, we use the term toxic to refer to intentional and even nonintentional usage of words, terms, or expressions that are frequently used in the context of offensive speech. Due to the high subjectivity for considering a comment toxic or not, but also the difficulty for most current chatbots to clearly distinguish intentional from nonintentional toxic comments, we prefer the usage of toxic as a more general term, while the track promotes the creation of chatbots that can be safely used across all environments and audiences (i.e., that can be used by large corporations and even for kids).

\subsection{Datasets and Baseline}
\label{subsec:datasets_baseline}
To allow participants to train and evaluate their models, we collected data from four different datasets. These datasets are preprocessed and formatted from their original sources as part of the Chat/Dialogue Modeling and Evaluation task (CHANEL) held during the 2020 Seventh Frederick Jelinek Memorial Summer Workshop\footnote{https://www.clsp.jhu.edu/workshops/20-workshop/}. The datasets are publicaly available at the CHANEL repository\footnote{https://github.com/CHANEL-JSALT-2020/datasets}. All the selected datasets are organized into turn of pairs (prompt-answer) and processed using Microsoft Azure Cognitive Services\footnote{https://azure.microsoft.com/en-us/services/cognitive-services/} to automatically detect toxic turns. Then, we select those pairs where the prompt was detected as toxic but the answer was not. To reduce false positives in the prompts or false negatives in the answers, we filter the Azure results by passing all detected turns through a dictionary consisting of 320 most common swear words in English. The dictionary is manually created from different lists on Internet including Wikipedia\footnote{https://en.wiktionary.org/wiki/Category:English\textunderscore swear\textunderscore words}, NoSwearing\footnote{https://www.noswearing.com/dictionary}, SlangDictionary\footnote{http://onlineslangdictionary.com/} and Hatebase\footnote{https://hatebase.org/}.
In concrete, the datasets we used are:
\subsubsection{MovieDic} Originally released by \cite{banchs2012movie}, this dataset consists of 65,215 dialogues (512k turns). The final selected set consists of 5,9k toxic pair turns.
\subsubsection{Cornell Movie Dataset} Originally released by \cite{danescu2011chameleons}, this dataset consists of 83,097 dialogues (304k turns). The final selected set consists of 3,2k turns.
\subsubsection{ChatCorpus} This dataset consists of dialogues from different datasets including movies, lyrics, and Twitter. We use the Twitter\_En\_Big dataset consisting of 754,5k turns)\footnote{https://github.com/Marsan-Ma/chat\_corpus/} organized into a single file where odd lines are considered the prompts and even lines are considered the answers. The final selected set consisted of 105,9k turns.
\subsubsection{DSTC8-Reddit} This dataset consists of 5,085,113 dialogues\footnote{https://github.com/microsoft/dstc8-reddit-corpus} collected from Reddit conversations and used during DSTC8 \cite{li2020results}. Our final selected set consists of 47,1k turns.
\hfill \break
\par
Besides the toxicity detection process, we extract additional features for the selected pairs to allow participants to apply additional filters for selecting data for training. In concrete, we remove entities (for annonymization purposes) using the Stanza library\footnote{https://stanfordnlp.github.io/stanza/ner.html}. We also extract humour score by using Colbert pretrained model\footnote{https://github.com/Moradnejad/ColBERT-Using-BERT-Sentence-Embedding-for-Humor-Detection} \cite{annamoradnejad2020colbert}, natural language inference for detecting entailment, contradiction and neutrality between prompts and answers\footnote{https://huggingface.co/microsoft/deberta-large-mnli} and sarcasm \footnote{https://huggingface.co/mrm8488/t5-base-finetuned-sarcasm-twitter}. 
Next, we perform emotion detection\footnote{https://drive.google.com/file/d/1lbEHWOFQt66n-T06cLYDbEhSmXlju4vx/view?usp=sharing} distinguishing up to 7 different emotions: happiness, sadness, fear, angry, surprise, disgust, and neutral \cite{cantelar2021genuine}. The use model is trained on four different datasets: Carer \cite{saravia2018carer}, DailyDialog \cite{li2017dailydialog}, EmpathicDialogs \cite{rashkin2018towards}, and EmotionLines \cite{chen2018emotionlines}. 
Finally, we also apply the Perspective API\footnote{https://developers.perspectiveapi.com/s/about-the-api} to detect the level of toxicity for both prompts and answers.
More detailed statistics for the four datasets are presented in Table~\ref{table_stats_datasets_task2}.

\begin{table*}[!ht]
\small
\ra{1.3}
\centering
\resizebox{\linewidth}{!}{
    \begin{tabular}{@{}lccc||ccc||ccc||ccc@{}}
    \toprule
    \multirow{2}{*}{} &     \multicolumn{3}{c}{MovieDic} & \multicolumn{3}{c}{Cornell} & \multicolumn{3}{c}{ChatCorpus} & \multicolumn{3}{c}{Reddit}  \\ 
    \cmidrule{2-4} \cmidrule{5-7} \cmidrule{8-10} \cmidrule{11-13}
        & Train & Dev & \multicolumn{1}{c}{Test} & Train & Dev & \multicolumn{1}{c}{Test} & Train & Dev & \multicolumn{1}{c}{Test} & Train & Dev & Test \\ \toprule
        No. Turns & 3359 & 720 & 1822 & 1829 & 392 & 995 & 74093 & 15877 & 15879 & 32977 & 7066 & 7067 \\ \hline
        Avg. turn length toxic & 16.3 & 16.3 & 20.6 & 17.3 & 17.5 & 15.3 & 15.5 & 15.5 & 15.5 & 24.1 & 24.1 & 24 \\ 
        Avg. turn length answer & 9.3 & 9.5 & 9.1 & 9.5 & 9.4 & 8.7 & 11.7 & 11.8 & 11.6 & 15.8 & 15.8 & 15.7 \\ \hline
        Avg. humour toxic & 0.95 & 0.94 & 0.96 & 0.95 & 0.95 & 0.92 & 0.92 & 0.92 & 0.92 & 0.95 & 0.95 & 0.95 \\ 
        Avg. humour answer & 0.78 & 0.78 & 0.78 & 0.79 & 0.8 & 0.77 & 0.81 & 0.81 & 0.8 & 0.85 & 0.85 & 0.84 \\ \hline
        Avg. sarcasm toxic & 0.53 & 0.55 & 0.51 & 0.53 & 0.53 & 0.53 & 0.61 & 0.62 & 0.61 & 0.62 & 0.63 & 0.61 \\ 
        Avg. sarcasm answer & 0.45 & 0.45 & 0.42 & 0.44 & 0.45 & 0.44 & 0.51 & 0.51 & 0.51 & 0.54 & 0.53 & 0.53 \\ \hline
        Avg. contradiction & 0.41 & 0.44 & 0.41 & 0.41 & 0.42 & 0.41 & 0.31 & 0.31 & 0.32 & 0.26 & 0.26 & 0.26 \\ 
        Avg. neutral & 0.55 & 0.51 & 0.55 & 0.55 & 0.53 & 0.54 & 0.66 & 0.66 & 0.66 & 0.72 & 0.71 & 0.72 \\ 
        Avg. entailment & 0.03 & 0.04 & 0.05 & 0.04 & 0.05 & 0.05 & 0.03 & 0.03 & 0.02 & 0.03 & 0.03 & 0.03 \\ \hline
        \multirow{2}{*}{Major emotion toxic} & A & A & A & A & A & N & A & A & A & A & A & A \\ 
& 38.7\% & 40.8\% & 36.3\% & 37.8\% & 37.0\% & 39.1\% & 35.2\% & 35.9\% & 35.7\% & 33.7\% & 33.9\% & 34.5\%
        \\
        \multirow{2}{*}{Major emotion answer} & N & N & N & N & N & N & H & H & H & N & N & N \\ 
& 62.4\% & 61.5\% & 57.1\% & 61.2\% & 65.0\% & 63.8\% & 29.3\% & 29.4\% & 29.7\% & 28.4\% & 14.9\% & 28.3\%        
        \\ \hline
        Avg. Perspective toxic & 0.79 & 0.8 & 0.79 & 0.77 & 0.77 & 0.65 & 0.81 & 0.8 & 0.81 & 0.8 & 0.8 & 0.8 \\ 
        Avg. Perspective answer & 0.15 & 0.16 & 0.15 & 0.15 & 0.14 & 0.14 & 0.22 & 0.22 & 0.22 & 0.16 & 0.16 & 0.16 \\ \bottomrule
    \end{tabular}
}
\caption{Statistics for the datasets used in Task 2. In the emotion rows, A, N and H mean Anger, Neutral and Happiness respectively.}
\label{table_stats_datasets_task2}
\end{table*}

Finally, participants are given as baseline a pretrained GPT-2 model trained on 147M multiturn dialogues from Reddit discussion threads \cite{zhang2019dialogpt} and finetuned on all our provided training data\footnote{\url{https://github.com/lfdharo/DSTC10_Track5_Toxicity}}.   

\subsection{Annotations}
\label{subsec:anotations}
To further assess the difficulty of the task, we manually annotate a subset of the test data. In total, 1290 prompt-answer pairs are annotated by 7 annotators from three different geographical zones (3 in the USA, 3 in Europe, and 1 in Asia). An annotation guideline, with no examples, is prepared to avoid biasing their responses. Only the toxic prompt is given as context and annotators are asked to annotate the answers in one of the following categories:

\subsubsection{Category ``1"}
1.	Any response that defuses toxicity or is a deflection.
2.	Any reasonable response that is non-committal. 
3.	Any response that a corporate chatbot could say with +80\% confidence if the exact or similar toxic utterance is given to it.
 
\subsubsection{Category ``0"}
1.	A response that a noncorporate or less restricted chatbot might say. 
2.	An answer that could be used after some minor manual fix or edition. 
3.	When the toxic comment is not clear, too general, or very domain specific, but the answer still could be used in general situations.

\subsubsection{Category ``-1"}
1.	In case that the prompt, the answer, or both are ungrammatical sentences requiring several manual editions to be considered relevant. 
2.	Answers that are overly general, therefore they are not engaging or they do not limit subsequent toxic behaviour.
3.	When neither the toxic comment nor the answer are good enough. 
4.	When either the prompt or answer utterances are too long.

\begin{figure}
    \centering
    \includegraphics[width=0.5\textwidth]{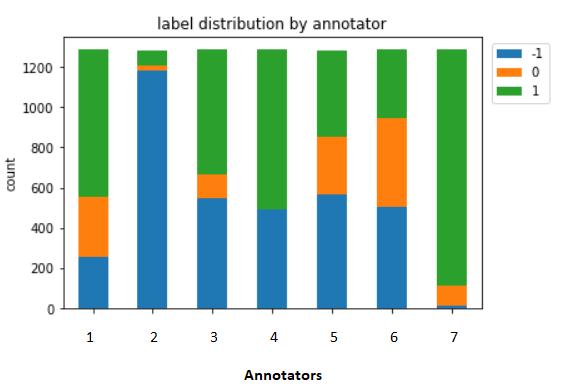}
    \caption{Label distribution between annotators}
    \label{fig:annotators_distribution}
\end{figure}

Then, we use the Fleiss' Kappa to measure the inter-annotator agreement. Unfortunately, the result is 0.1 which, after deeper analysis and discussion, is attributed to cultural differences between the annotators (i.e., differences in the consideration that something is toxic or not due to specific swearing words, intention or usability of the answers) as it has been pointed also by \cite{leonardelli2021agreeing}. Figure \ref{fig:annotators_distribution} shows the label distribution among annotators. 

\begin{table}
\centering
    \begin{tabular}{@{}cccccc@{}}
    \toprule
     & 3 & 4 & 5 & 6 & 7\\
    \toprule
    -1 & 57 & 220 & 164 & 62 & 0 \\
    0 & 59 & 4 & 0 & 0 & 0 \\
    1 & 192 & 232 & 176 & 102 & 19 \\ \bottomrule
    \end{tabular}
\caption{Statistics for the human annotation of 1290 prompts-answers turns.}
\label{table_stats_annotations_task2}
\end{table}

\subsection{Results}
As there was no submission in this subtask, we decided to perform some objective and human evaluations by taking a selected set of toxic-answer prompts as described below.

\subsubsection{Data preparation}
Table \ref{table_stats_annotations_task2} shows the number of turn pairs where more than 3 annotators agreed on the three possible labels (-1, 0, and 1) as described in section \ref{subsec:anotations}. Based on these statistics, and with the purpose of comparing the output of different existing SotA chatbot models, we selected a subset of pairs where more than 5 annotators agreed. In this case, for  label 1 we selected a total of 297 turns, while for label -1 we selected a total of 62. 

The toxic prompts for the 359 selected turns were used as seeds to three different pretrained models: a) the pretrained baseline released to the participants (see section \ref{subsec:datasets_baseline} based on a finetuned version of DialogGPT,  b) BlenderBot vs 2.0 including its safety layer \cite{xu2021beyond,komeili2021internet}, and c) GPT-3 \cite{brown2020language} available using the OpenAI API\footnote{https://beta.openai.com/?app=chat} and obtaining the response of the DaVinci version using as input prompt ``The following is a conversation with an AI assistant. The assistant is helpful, creative, clever, and very friendly", followed by the toxic prompt and saving the provided answer.

\subsubsection{Subjective Metrics} In this case, we performed manual annotations on a subset of the 359 selected turns from test and performed the binary task of assessing the quality of a pair of system answers given the toxic prompt. Below we describe the process in detail.

From the 359 sentences, we randomly selected 160 toxic prompts and created all possible pair combinations from the three possible chatbot responses. Then, we asked 7 annotators to perform the binary task of indicating which system was providing a better answer to the given toxic prompt. To avoid any bias, we randomly distributed the answers given by any of the selected chatbots. In addition, we asked the annotators to indicate whether the answer provided by any or both chatbots was also toxic or could promote the user's misbehavior. Finally, and as a control measure, we added 60 random pairs where the original human answer was compared against the three selected chatbots. Therefore, the total number of annotated items per annotator was 510 pairs.  

A guideline was given to the annotators indicating to analyze the toxic prompt against the two possible answers and then selecting among these three options: a) A or B: to select the winner system, b) T (for tied): in case both answers were good, and c) U (for unrelated): in case both answers were completely unrelated to the prompt, wrong or unnatural. Moreover, we asked annotators to flag any of the answers in case they contain toxicity or promote the user's misbehavior.  
Table \ref{table_stats_performance_baselines_task2} shows the statistics of the annotation where we compare the number of times a given system was selected over others, as well as the number of times it was not selected, or its response tied with another, or was unrelated/bad to the given prompt. Take into account that for human statistics only 60 sentences per annotator were annotated. Percentages are calculated over the total number of items annotated. 

\begin{figure}
    \centering
    \includegraphics[width=0.5\textwidth]{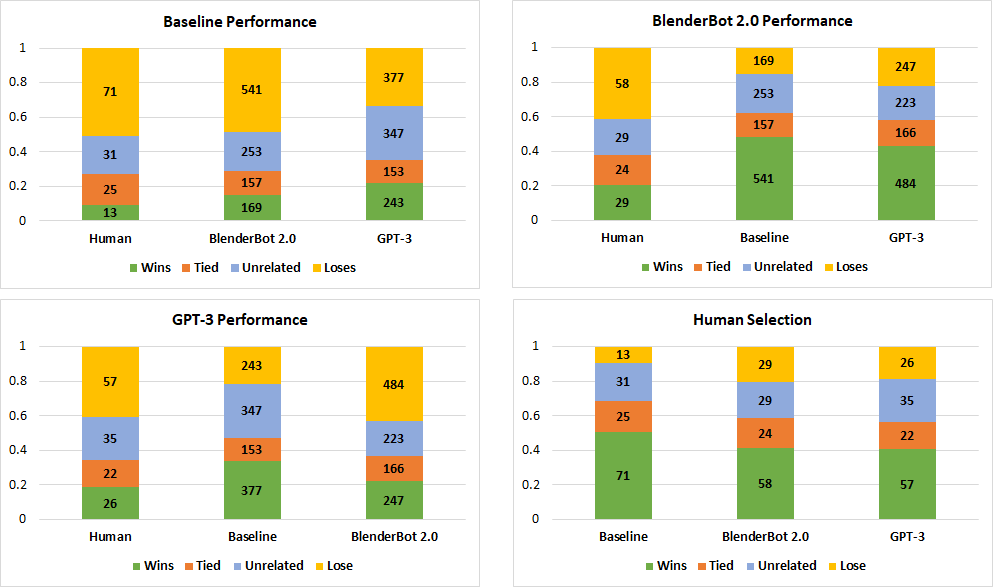}
    \caption{Comparative performance between the different chatbots and human answers on the annotated test set}
    \label{fig:comparison_chatbots_task2}
\end{figure}

From the table, we can see that BlenderBot vs 2.0 performs the best (i.e., wins 44.3\% of the times and with the same result as the original human answer) when compared with the other options, while GPT-3 is selected in the second place (27.3\% of the times). The baseline is third with 17.9\% of its answer being unrelated. 
Surprisingly, human answers are not always selected (i.e., they lose 16.2\% of the time) and even they can be as good as other chatbots answers  16.9\% of the time. In addition,  human answers are considered not good (i.e., unrelated) 22.6\% of the time which is a similar percentage obtained by the other chatbots. Refer to figure \ref{fig:comparison_chatbots_task2} for detailed information about the performance of each chatbot in comparison with the others or the human answer. 

On the other hand, the results about how many times a response given by a chatbot was flagged (i.e., containing toxic or not engaging answers) show that our baseline was flagged 13.8\% of the time, BlenderBot vs 2.0 9.9\%, GPT-3 at 14.9\%, and human answers were flagged 7.8\% of the time. 

These results probe the difficulty of the task due to the lack of context and how difficult it is to provide answers that are simultaneously informative, engaging, and nontoxic.

\begin{table}[!ht]	
    \centering
    \resizebox{\linewidth}{!}{
    \begin{tabular}{ccccc}
    \toprule
        ~ & Wins & Tied & Unrelated & Loses \\ \toprule
        \multirow{2}{*}{Baseline} & 425 & 335 & 631 & 989 \\
                 & 17.9\% & 14.1\% & 26.5\% & 41.6\% \\
        \cmidrule{2-5}
        \multirow{2}{*}{BlenderBot Vs 2.0} & 1054 & 347 & 505 & 474\\
        & 44.3\% & 14.6\% & 21.2\% & 19.9\% \\
        \cmidrule{2-5}
        \multirow{2}{*}{GPT-3} & 650 & 341 & 605 & 784 \\ 
        & 27.3\% & 14.3\% & 25.4\% & 32.9\% \\
        \cmidrule{2-5}
        \multirow{2}{*}{Human} & 186 & 71 & 95 & 68 \\
        & 44.3\% & 16.9\% & 22.6\% & 16.2\% \\
        \hline \hline
        Total & 2315 & 1094 & 1836 & 2315 \\
        \bottomrule
    \end{tabular}
    }
    \caption{Comparison of the human selections for the test set on subtask 2. Percentage are over  total no. annotated items for each chatbot.}
    \label{table_stats_performance_baselines_task2}    
\end{table}

\subsubsection{Objective Metrics} In this case, we took the same 359 sentences selected from the test set and the three generated chatbot outputs, comparing the outputs with the original ground truth (i.e., human answers) through four different objective metrics: a) BLEU: proposed by \cite{papineni2002bleu}, and more specifically through the SacreBleu implementation \cite{post-2018-call} providing scores from 0.0 to 1.0, b) ROUGE: proposed by \cite{lin-2004-rouge}, which calculates the longest common subsequence (LCS)  between each pair of reference and candidate sentences providing scores  c) BERT-Score: proposed by \cite{bert-score} which uses pretrained contextual embeddings from BERT and then matching words in both the candidate and reference sentences by cosine similarity (i.e., provide values scores between 0.0 and 1.0), and d) BLEURT: proposed by \cite{bleurt}, a BERT-based model pretrained on synthetic data and fine-tuned on human annotations; in this model, positive scores are related with better responses.

Table \ref{table_objective_baselines_task2} shows the results obtained for each of the evaluated chatbots. The results for the word-matching metrics (BLEU and ROUGE) are very low due to the high differences in the chatbot generated responses and the human ones, which does not necessarily mean they are bad, but
syntactically different. On the other hand, semantic metrics (i.e., BERTScore and BLEURT) show marginal differences between chatbots, with a slight advantage for BlenderBot.  

\begin{table}
    \small
    \centering
    \resizebox{\linewidth}{!}{
    \begin{tabular}{ccccc}
        \toprule
        System & BLEU & ROUGE & BERTScore & BLEURT \\ 
        \toprule
        Baseline & 0.008 & 0.072 & 0.832 & -1.180 \\ 
        BlenderBot Vs 2.0 & 0.009 & 0.097 & 0.836 & -1.183 \\
        GPT-3 & 0.008 & 0.065 & 0.831 & -1.201 \\
        \bottomrule
    \end{tabular}
    }
    \caption{Objective metrics for tested chatbots in subtask 2.}
    \label{table_objective_baselines_task2}     
\end{table}

\section{Conclusions and Future Work}
This paper presents a comprehensive overview of Track 5 on ``Automatic Evaluation and Moderation of Open-domain Dialogue Systems" organized as part of the 10\textsuperscript{th} Dialogue System Technology Challenge (DSTC10). The track was organized in two subtasks aimed at addressing two important problems of the state-of-the-art in Dialogue Systems: the design of automatic evaluation metrics to propel the research and development cycles of dialogue technologies, and the management and moderation of offensive and toxic interactions to increase the safety of conversational systems. 

The first subtask included active participation from nine teams, resulting in interesting contributions to the state-of-the-art on the specific problem of automatic evaluation of chat-oriented dialogue systems; however, these still seems to be room for significant improvements. The subtask assessed the performance of the submitted evaluation metrics against a reference-free deep AMFM baseline \cite{zhang-etal-2021-deepamfm} over a collection of 19 different chatbot datasets (14 development and 5 test).

The second subtask, focused on the moderation of dialogue systems, ended up without submissions. However, the organizing team managed to propose and evaluate three different baseline systems, setting up a reference framework for an eventual rerun of the subtask in future editions of DSTC or similar venues. The management and moderation of offensive and toxic interactions is a nascent area of research of fundamental importance for ensuring the development of safe conversational system technologies.

As future work, we plan to continue increasing the coverage of the current datasets, as well as improving the baseline systems to make both challenge subtasks more competitive and attract new participants to the corresponding future editions. In detail, for subtask 1 we plan to include the unification of dimensions across existing datasets, the generation of annotations at dialogue level, and the incorporation of new dimensions like toxicity and bias. As for subtask 2, we plan to include splitting the subtask into two parts: classification of the toxic comment and controlled generation based on the detected toxicity type.

\label{section:conclusions} 

\section*{Acknowledgments}
We want to thank Mario Rodríguez-Cantelar and Marcos Estecha for their contribution to the annotation of data for subtask 2. We want to thank the JSALT2020 organizers: Sanjeev Khudanpur, Najim Dehak, Jan Trmal, and Piotr Żelasko, as well as the JSALT2020 workshop sponsors: Amazon and Microsoft, and Ms Azure services (especially Irving Kwong) for their support to collect, process, and automatically annotate the datasets used in subtask 2.

\bibliography{references}

\begin{thebibliography}{55}
\providecommand{\natexlab}[1]{#1}

\bibitem[{Adiwardana et~al.(2020)Adiwardana, Luong, So, Hall, Fiedel,
  Thoppilan, Yang, Kulshreshtha, Nemade, Lu et~al.}]{adiwardana2020towards}
Adiwardana, D.; Luong, M.-T.; So, D.~R.; Hall, J.; Fiedel, N.; Thoppilan, R.;
  Yang, Z.; Kulshreshtha, A.; Nemade, G.; Lu, Y.; et~al. 2020.
\newblock Towards a human-like open-domain chatbot.
\newblock \emph{arXiv preprint arXiv:2001.09977}.

\bibitem[{Annamoradnejad and Zoghi(2020)}]{annamoradnejad2020colbert}
Annamoradnejad, I.; and Zoghi, G. 2020.
\newblock Colbert: Using bert sentence embedding for humor detection.
\newblock \emph{arXiv preprint arXiv:2004.12765}.

\bibitem[{Banchs(2012)}]{banchs2012movie}
Banchs, R.~E. 2012.
\newblock Movie-DiC: a movie dialogue corpus for research and development.
\newblock In \emph{Proceedings of the 50th Annual Meeting of ACL (Volume 2:
  Short Papers)}, 203--207.

\bibitem[{Brown et~al.(2020)Brown, Mann, Ryder, Subbiah, Kaplan, Dhariwal,
  Neelakantan, Shyam, Sastry, Askell et~al.}]{brown2020language}
Brown, T.~B.; Mann, B.; Ryder, N.; Subbiah, M.; Kaplan, J.; Dhariwal, P.;
  Neelakantan, A.; Shyam, P.; Sastry, G.; Askell, A.; et~al. 2020.
\newblock Language models are few-shot learners.
\newblock \emph{arXiv preprint arXiv:2005.14165}.

\bibitem[{Chen et~al.(2018)Chen, Hsu, Kuo, Ku et~al.}]{chen2018emotionlines}
Chen, S.-Y.; Hsu, C.-C.; Kuo, C.-C.; Ku, L.-W.; et~al. 2018.
\newblock Emotionlines: An emotion corpus of multi-party conversations.
\newblock \emph{arXiv preprint arXiv:1802.08379}.

\bibitem[{Danescu-Niculescu-Mizil and Lee(2011)}]{danescu2011chameleons}
Danescu-Niculescu-Mizil, C.; and Lee, L. 2011.
\newblock Chameleons in imagined conversations: A new approach to understanding
  coordination of linguistic style in dialogs.
\newblock \emph{arXiv preprint arXiv:1106.3077}.

\bibitem[{Dinan et~al.(2020)Dinan, Logacheva, Malykh, Miller, Shuster, Urbanek,
  Kiela, Szlam, Serban, Lowe et~al.}]{dinan2020second}
Dinan, E.; Logacheva, V.; Malykh, V.; Miller, A.; Shuster, K.; Urbanek, J.;
  Kiela, D.; Szlam, A.; Serban, I.; Lowe, R.; et~al. 2020.
\newblock The second conversational intelligence challenge (convai2).
\newblock In \emph{The NeurIPS'18 Competition}, 187--208. Springer.

\bibitem[{Fortuna, Soler, and Wanner(2020)}]{fortuna2020toxic}
Fortuna, P.; Soler, J.; and Wanner, L. 2020.
\newblock Toxic, hateful, offensive or abusive? what are we really classifying?
  an empirical analysis of hate speech datasets.
\newblock In \emph{Proceedings of the 12th language resources and evaluation
  conference}, 6786--6794.

\bibitem[{Galley et~al.(2019)Galley, Brockett, Gao, Gao, and
  Dolan}]{Galley2019GroundedRG}
Galley, M.; Brockett, C.; Gao, X.; Gao, J.; and Dolan, B. 2019.
\newblock Grounded Response Generation Task at DSTC7.
\newblock In \emph{AAAI Dialog System Technology Challenges Workshop}.

\bibitem[{Giorgi, Ungar, and Schwartz(2021)}]{giorgi2021characterizing}
Giorgi, S.; Ungar, L.; and Schwartz, H.~A. 2021.
\newblock Characterizing Social Spambots by their Human Traits.
\newblock In \emph{Findings of ACL: ACL-IJCNLP 2021}, 5148--5158. Online:
  Association for Computational Linguistics.

\bibitem[{Gopalakrishnan et~al.(2019)Gopalakrishnan, Hedayatnia, Chen,
  Gottardi, Kwatra, Venkatesh, Gabriel, Hakkani-T{\"u}r, and
  AI}]{gopalakrishnan2019topical}
Gopalakrishnan, K.; Hedayatnia, B.; Chen, Q.; Gottardi, A.; Kwatra, S.;
  Venkatesh, A.; Gabriel, R.; Hakkani-T{\"u}r, D.; and AI, A.~A. 2019.
\newblock Topical-Chat: Towards Knowledge-Grounded Open-Domain Conversations.
\newblock In \emph{INTERSPEECH}, 1891--1895.

\bibitem[{Gupta et~al.(2019)Gupta, Mehri, Zhao, Pavel, Eskenazi, and
  Bigham}]{gupta-etal-2019-investigating}
Gupta, P.; Mehri, S.; Zhao, T.; Pavel, A.; Eskenazi, M.; and Bigham, J. 2019.
\newblock Investigating Evaluation of Open-Domain Dialogue Systems With Human
  Generated Multiple References.
\newblock In \emph{Proceedings of the 20th Annual SIGdial Meeting on Discourse
  and Dialogue}, 379--391. Stockholm, Sweden: Association for Computational
  Linguistics.

\bibitem[{Hori and Hori(2017)}]{hori2017end}
Hori, C.; and Hori, T. 2017.
\newblock End-to-end conversation modeling track in DSTC6.
\newblock \emph{arXiv preprint arXiv:1706.07440}.

\bibitem[{Huang et~al.(2020)Huang, Ye, Qin, Lin, and
  Liang}]{huang-etal-2020-grade}
Huang, L.; Ye, Z.; Qin, J.; Lin, L.; and Liang, X. 2020.
\newblock {GRADE}: Automatic Graph-Enhanced Coherence Metric for Evaluating
  Open-Domain Dialogue Systems.
\newblock In \emph{Proceedings of the 2020 Conference on Empirical Methods in
  Natural Language Processing (EMNLP)}.

\bibitem[{Komeili, Shuster, and Weston(2021)}]{komeili2021internet}
Komeili, M.; Shuster, K.; and Weston, J. 2021.
\newblock Internet-augmented dialogue generation.
\newblock \emph{arXiv preprint arXiv:2107.07566}.

\bibitem[{Kong-Vega et~al.(2019)Kong-Vega, Shen, Wang, and
  D’Haro}]{kong2019subjective}
Kong-Vega, N.; Shen, M.; Wang, M.; and D’Haro, L.~F. 2019.
\newblock Subjective annotation and evaluation of three different chatbots
  WOCHAT: shared task report.
\newblock In \emph{9th International Workshop on Spoken Dialogue System
  Technology}, 371--378. Springer.

\bibitem[{Lee, Lim, and Sedoc(2020)}]{lee2020evaluation}
Lee, S.; Lim, H.; and Sedoc, J. 2020.
\newblock An Evaluation Protocol for Generative Conversational Systems.
\newblock \emph{arXiv preprint arXiv:2010.12741}.

\bibitem[{Leonardelli et~al.(2021)Leonardelli, Menini, Aprosio, Guerini, and
  Tonelli}]{leonardelli2021agreeing}
Leonardelli, E.; Menini, S.; Aprosio, A.~P.; Guerini, M.; and Tonelli, S. 2021.
\newblock Agreeing to Disagree: Annotating Offensive Language Datasets with
  Annotators' Disagreement.
\newblock In \emph{Proceedings of the 2021 Conference on Empirical Methods in
  Natural Language Processing}, 10528--10539.

\bibitem[{Li et~al.(2020)Li, Peng, Lee, Gao, Takanobu, Zhu, Huang, Schulz,
  Atkinson, and Adada}]{li2020results}
Li, J.; Peng, B.; Lee, S.; Gao, J.; Takanobu, R.; Zhu, Q.; Huang, M.; Schulz,
  H.; Atkinson, A.; and Adada, M. 2020.
\newblock Results of the multi-domain task-completion dialog challenge.
\newblock In \emph{Proceedings of the 34th AAAI Conference on Artificial
  Intelligence, Eighth Dialog System Technology Challenge Workshop}.

\bibitem[{Li et~al.(2017{\natexlab{a}})Li, Su, Shen, Li, Cao, and
  Niu}]{li2017dailydialog}
Li, Y.; Su, H.; Shen, X.; Li, W.; Cao, Z.; and Niu, S. 2017{\natexlab{a}}.
\newblock Dailydialog: A manually labelled multi-turn dialogue dataset.
\newblock \emph{arXiv preprint arXiv:1710.03957}.

\bibitem[{Li et~al.(2017{\natexlab{b}})Li, Su, Shen, Li, Cao, and
  Niu}]{li-etal-2017-dailydialog}
Li, Y.; Su, H.; Shen, X.; Li, W.; Cao, Z.; and Niu, S. 2017{\natexlab{b}}.
\newblock {D}aily{D}ialog: A Manually Labelled Multi-turn Dialogue Dataset.
\newblock In \emph{Proceedings of the Eighth International Joint Conference on
  Natural Language Processing}, 986--995.

\bibitem[{Lin(2004)}]{lin-2004-rouge}
Lin, C.-Y. 2004.
\newblock {ROUGE}: A Package for Automatic Evaluation of Summaries.
\newblock In \emph{Text Summarization Branches Out}, 74--81. Barcelona, Spain:
  Association for Computational Linguistics.

\bibitem[{Liu et~al.(2016)Liu, Lowe, Serban, Noseworthy, Charlin, and
  Pineau}]{liu2016not}
Liu, C.-W.; Lowe, R.; Serban, I.~V.; Noseworthy, M.; Charlin, L.; and Pineau,
  J. 2016.
\newblock How not to evaluate your dialogue system: An empirical study of
  unsupervised evaluation metrics for dialogue response generation.
\newblock \emph{arXiv preprint arXiv:1603.08023}.

\bibitem[{Liu et~al.(2019)Liu, Ott, Goyal, Du, Joshi, Chen, Levy, Lewis,
  Zettlemoyer, and Stoyanov}]{liu2019roberta}
Liu, Y.; Ott, M.; Goyal, N.; Du, J.; Joshi, M.; Chen, D.; Levy, O.; Lewis, M.;
  Zettlemoyer, L.; and Stoyanov, V. 2019.
\newblock {R}o{BERT}a: A Robustly Optimized BERT Pretraining Approach.
\newblock \emph{arXiv preprint arXiv:1907.11692}.

\bibitem[{Mehri and
  Eskenazi(2020{\natexlab{a}})}]{mehri-eskenazi-2020-unsupervised}
Mehri, S.; and Eskenazi, M. 2020{\natexlab{a}}.
\newblock Unsupervised Evaluation of Interactive Dialog with {D}ialo{GPT}.
\newblock In \emph{Proceedings of the 21th Annual Meeting of the SigDial},
  225--235.

\bibitem[{Mehri and Eskenazi(2020{\natexlab{b}})}]{mehri-eskenazi-2020-usr}
Mehri, S.; and Eskenazi, M. 2020{\natexlab{b}}.
\newblock {USR}: An Unsupervised and Reference Free Evaluation Metric for
  Dialog Generation.
\newblock In \emph{Proceedings of the 58th Annual Meeting of ACL}.

\bibitem[{Merdivan et~al.(2020)Merdivan, Singh, Hanke, Kropf, Holzinger, and
  Geist}]{app10030762}
Merdivan, E.; Singh, D.; Hanke, S.; Kropf, J.; Holzinger, A.; and Geist, M.
  2020.
\newblock Human Annotated Dialogues Dataset for Natural Conversational Agents.
\newblock \emph{Applied Sciences}, 10.

\bibitem[{Pang et~al.(2020)Pang, Nijkamp, Han, Zhou, Liu, and
  Tu}]{pang-etal-2020-towards}
Pang, B.; Nijkamp, E.; Han, W.; Zhou, L.; Liu, Y.; and Tu, K. 2020.
\newblock Towards Holistic and Automatic Evaluation of Open-Domain Dialogue
  Generation.
\newblock In \emph{Proceedings of the 58th Annual Meeting of ACL}, 3619--3629.
  Online: Association for Computational Linguistics.

\bibitem[{Papineni et~al.(2002)Papineni, Roukos, Ward, and
  Zhu}]{papineni2002bleu}
Papineni, K.; Roukos, S.; Ward, T.; and Zhu, W.-J. 2002.
\newblock Bleu: a method for automatic evaluation of machine translation.
\newblock In \emph{Proceedings of the 40th annual meeting of ACL}, 311--318.

\bibitem[{Phy, Zhao, and Aizawa(2020)}]{phy-etal-2020-deconstruct}
Phy, V.; Zhao, Y.; and Aizawa, A. 2020.
\newblock Deconstruct to Reconstruct a Configurable Evaluation Metric for
  Open-Domain Dialogue Systems.
\newblock In \emph{Proceedings of the 28th International Conference on
  Computational Linguistics}, 4164--4178. Barcelona, Spain (Online):
  International Committee on Computational Linguistics.

\bibitem[{Post(2018)}]{post-2018-call}
Post, M. 2018.
\newblock A Call for Clarity in Reporting {BLEU} Scores.
\newblock In \emph{Proceedings of the Third Conference on Machine Translation:
  Research Papers}, 186--191. Belgium, Brussels: Association for Computational
  Linguistics.

\bibitem[{Radford et~al.(2019)Radford, Wu, Child, Luan, Amodei, Sutskever
  et~al.}]{gpt2}
Radford, A.; Wu, J.; Child, R.; Luan, D.; Amodei, D.; Sutskever, I.; et~al.
  2019.
\newblock Language models are unsupervised multitask learners.
\newblock \emph{OpenAI blog}.

\bibitem[{Raffel et~al.(2020)Raffel, Shazeer, Roberts, Lee, Narang, Matena,
  Zhou, Li, and Liu}]{raffel2020exploring}
Raffel, C.; Shazeer, N.; Roberts, A.; Lee, K.; Narang, S.; Matena, M.; Zhou,
  Y.; Li, W.; and Liu, P.~J. 2020.
\newblock Exploring the Limits of Transfer Learning with a Unified Text-to-Text
  Transformer.
\newblock \emph{Journal of Machine Learning Research}, 21(140): 1--67.

\bibitem[{Rashkin et~al.(2018)Rashkin, Smith, Li, and
  Boureau}]{rashkin2018towards}
Rashkin, H.; Smith, E.~M.; Li, M.; and Boureau, Y.-L. 2018.
\newblock Towards empathetic open-domain conversation models: A new benchmark
  and dataset.
\newblock \emph{arXiv preprint arXiv:1811.00207}.

\bibitem[{Rashkin et~al.(2019)Rashkin, Smith, Li, and
  Boureau}]{rashkin-etal-2019-towards}
Rashkin, H.; Smith, E.~M.; Li, M.; and Boureau, Y.-L. 2019.
\newblock Towards Empathetic Open-domain Conversation Models: A New Benchmark
  and Dataset.
\newblock In \emph{Proceedings of the 57th Annual Meeting of ACL}, 5370--5381.

\bibitem[{Rodríguez-Cantelar et~al.(2021)Rodríguez-Cantelar, de~la Cal,
  Estecha, Gutiérrez, Martín, Milara, Jiménez, and
  D’Haro}]{cantelar2021genuine}
Rodríguez-Cantelar, M.; de~la Cal, D.; Estecha, M.; Gutiérrez, A.~G.;
  Martín, D.; Milara, N. R.~N.; Jiménez, R.~M.; and D’Haro, L.~F. 2021.
\newblock Genuine\textsuperscript{2}: An open domain chatbot based on
  generative models.
\newblock \emph{Proceedings Alexa Socialbot Grand Challenge SGC4}.

\bibitem[{Roller et~al.(2021)Roller, Dinan, Goyal, Ju, Williamson, Liu, Xu,
  Ott, Smith, Boureau, and Weston}]{roller-etal-2021-recipes}
Roller, S.; Dinan, E.; Goyal, N.; Ju, D.; Williamson, M.; Liu, Y.; Xu, J.; Ott,
  M.; Smith, E.~M.; Boureau, Y.-L.; and Weston, J. 2021.
\newblock Recipes for Building an Open-Domain Chatbot.
\newblock In \emph{Proceedings of the 16th Conference of the European Chapter
  of ACL: Main Volume}, 300--325. Online: Association for Computational
  Linguistics.

\bibitem[{Sai et~al.(2020)Sai, Mohankumar, Arora, and
  Khapra}]{DBLP:journals/tacl/SaiMAK20}
Sai, A.~B.; Mohankumar, A.~K.; Arora, S.; and Khapra, M.~M. 2020.
\newblock Improving Dialog Evaluation with a Multi-reference Adversarial
  Dataset and Large Scale Pretraining.
\newblock \emph{Trans. Assoc. Comput. Linguistics}, 8: 810--827.

\bibitem[{Saravia et~al.(2018)Saravia, Liu, Huang, Wu, and
  Chen}]{saravia2018carer}
Saravia, E.; Liu, H.-C.~T.; Huang, Y.-H.; Wu, J.; and Chen, Y.-S. 2018.
\newblock Carer: Contextualized affect representations for emotion recognition.
\newblock In \emph{Proceedings of the 2018 Conference on Empirical Methods in
  Natural Language Processing}, 3687--3697.

\bibitem[{Sedoc et~al.(2019)Sedoc, Ippolito, Kirubarajan, Thirani, Ungar, and
  Callison-Burch}]{sedoc-etal-2019-chateval}
Sedoc, J.; Ippolito, D.; Kirubarajan, A.; Thirani, J.; Ungar, L.; and
  Callison-Burch, C. 2019.
\newblock {C}hat{E}val: A Tool for Chatbot Evaluation.
\newblock In \emph{Proceedings of the 2019 Conference of the North {A}merican
  Chapter of ACL (Demonstrations)}, 60--65. Minneapolis, Minnesota: Association
  for Computational Linguistics.

\bibitem[{See et~al.(2019)See, Roller, Kiela, and Weston}]{see-etal-2019-makes}
See, A.; Roller, S.; Kiela, D.; and Weston, J. 2019.
\newblock What makes a good conversation? How controllable attributes affect
  human judgments.
\newblock In \emph{Proceedings of the 2019 Conference of the North {A}merican
  Chapter of ACL: Human Language Technologies, Volume 1 (Long and Short
  Papers)}, 1702--1723. Minneapolis, Minnesota: Association for Computational
  Linguistics.

\bibitem[{Sellam, Das, and Parikh(2020)}]{bleurt}
Sellam, T.; Das, D.; and Parikh, A.~P. 2020.
\newblock BLEURT: Learning Robust Metrics for Text Generation.
\newblock In \emph{ACL}.

\bibitem[{Smith et~al.(2020)Smith, Williamson, Shuster, Weston, and
  Boureau}]{smith2020can}
Smith, E.~M.; Williamson, M.; Shuster, K.; Weston, J.; and Boureau, Y.-L. 2020.
\newblock Can you put it all together: Evaluating conversational agents'
  ability to blend skills.
\newblock \emph{arXiv preprint arXiv:2004.08449}.

\bibitem[{Vinyals and Le(2015)}]{vinyals2015neural}
Vinyals, O.; and Le, Q. 2015.
\newblock A neural conversational model.
\newblock \emph{arXiv preprint arXiv:1506.05869}.

\bibitem[{Xu et~al.(2020)Xu, Ju, Li, Boureau, Weston, and
  Dinan}]{xu2020recipes}
Xu, J.; Ju, D.; Li, M.; Boureau, Y.-L.; Weston, J.; and Dinan, E. 2020.
\newblock Recipes for safety in open-domain chatbots.
\newblock \emph{arXiv preprint arXiv:2010.07079}.

\bibitem[{Xu, Szlam, and Weston(2021)}]{xu2021beyond}
Xu, J.; Szlam, A.; and Weston, J. 2021.
\newblock Beyond goldfish memory: Long-term open-domain conversation.
\newblock \emph{arXiv preprint arXiv:2107.07567}.

\bibitem[{Ye et~al.(2021)Ye, Lu, Huang, Lin, and
  Liang}]{DBLP:conf/acl/YeLHLL20}
Ye, Z.; Lu, L.; Huang, L.; Lin, L.; and Liang, X. 2021.
\newblock Towards Quantifiable Dialogue Coherence Evaluation.
\newblock In Zong, C.; Xia, F.; Li, W.; and Navigli, R., eds.,
  \emph{Proceedings of the 59th Annual Meeting of ACL and 11th International
  Joint Conference on Natural Language Processing, {ACL/IJCNLP} 2021, (Volume
  1: Long Papers), August 1-6, 2021}, 2718--2729. Association for Computational
  Linguistics.

\bibitem[{Yeh, Eskenazi, and Mehri(2021)}]{yeh2021comprehensive}
Yeh, Y.-T.; Eskenazi, M.; and Mehri, S. 2021.
\newblock A Comprehensive Assessment of Dialog Evaluation Metrics.
\newblock \emph{arXiv preprint arXiv:2106.03706}.

\bibitem[{Zhang et~al.(2021{\natexlab{a}})Zhang, Chen, D'Haro, Zhang,
  Friedrichs, Lee, and Li}]{DBLP:conf/acl/ZhangCDZFL020}
Zhang, C.; Chen, Y.; D'Haro, L.~F.; Zhang, Y.; Friedrichs, T.; Lee, G.; and Li,
  H. 2021{\natexlab{a}}.
\newblock DynaEval: Unifying Turn and Dialogue Level Evaluation.
\newblock In Zong, C.; Xia, F.; Li, W.; and Navigli, R., eds.,
  \emph{Proceedings of the 59th Annual Meeting of ACL and the 11th
  International Joint Conference on Natural Language Processing, {ACL/IJCNLP}
  2021, (Volume 1: Long Papers), Virtual Event, August 1-6, 2021}, 5676--5689.
  Association for Computational Linguistics.

\bibitem[{Zhang et~al.(2021{\natexlab{b}})Zhang, D'Haro, Banchs, Friedrichs,
  and Li}]{zhang-etal-2021-deepamfm}
Zhang, C.; D'Haro, L.~F.; Banchs, R.~E.; Friedrichs, T.; and Li, H.
  2021{\natexlab{b}}.
\newblock \emph{Deep AM-FM: Toolkit for Automatic Dialogue Evaluation}, 53--69.
\newblock Singapore: Springer Singapore.
\newblock ISBN 978-981-15-8395-7.

\bibitem[{Zhang et~al.(2021{\natexlab{c}})Zhang, D'Haro, Chen, Friedrichs, and
  Li}]{zhang2021investigating}
Zhang, C.; D'Haro, L.~F.; Chen, Y.; Friedrichs, T.; and Li, H.
  2021{\natexlab{c}}.
\newblock Investigating the Impact of Pre-trained Language Models on Dialog
  Evaluation.
\newblock \emph{arXiv preprint arXiv:2110.01895}.

\bibitem[{Zhang et~al.(2018)Zhang, Dinan, Urbanek, Szlam, Kiela, and
  Weston}]{zhang2018personalizing}
Zhang, S.; Dinan, E.; Urbanek, J.; Szlam, A.; Kiela, D.; and Weston, J. 2018.
\newblock Personalizing dialogue agents: I have a dog, do you have pets too?
\newblock \emph{arXiv preprint arXiv:1801.07243}.

\bibitem[{Zhang* et~al.(2020)Zhang*, Kishore*, Wu*, Weinberger, and
  Artzi}]{bert-score}
Zhang*, T.; Kishore*, V.; Wu*, F.; Weinberger, K.~Q.; and Artzi, Y. 2020.
\newblock BERTScore: Evaluating Text Generation with BERT.
\newblock In \emph{International Conference on Learning Representations}.

\bibitem[{Zhang et~al.(2019)Zhang, Sun, Galley, Chen, Brockett, Gao, Gao, Liu,
  and Dolan}]{zhang2019dialogpt}
Zhang, Y.; Sun, S.; Galley, M.; Chen, Y.-C.; Brockett, C.; Gao, X.; Gao, J.;
  Liu, J.; and Dolan, B. 2019.
\newblock Dialo{GPT}: Large-scale generative pre-training for conversational
  response generation.
\newblock \emph{arXiv preprint arXiv:1911.00536}.

\bibitem[{Zhao, Lala, and Kawahara(2020)}]{zhao-etal-2020-designing}
Zhao, T.; Lala, D.; and Kawahara, T. 2020.
\newblock Designing Precise and Robust Dialogue Response Evaluators.
\newblock In \emph{Proceedings of the 58th Annual Meeting of ACL}, 26--33.

\end{thebibliography}

\end{document}